\documentclass{article} 
\usepackage{iclr2026_conference,times}

\usepackage{amsmath,amsfonts,bm}

\def\eqref#1{equation~\ref{#1}}

\def\1{\bm{1}}

\DeclareMathAlphabet{\mathsfit}{\encodingdefault}{\sfdefault}{m}{sl}
\SetMathAlphabet{\mathsfit}{bold}{\encodingdefault}{\sfdefault}{bx}{n}

\usepackage{hyperref}
\usepackage{url}
\usepackage{multirow}
\usepackage{multicol}
\usepackage{makecell}
\usepackage{algorithm}
\usepackage{algpseudocode}
\usepackage{subcaption}
\usepackage{graphicx} 
\usepackage{booktabs}
\usepackage{caption}
\usepackage{wrapfig}

\title{Learning What to Trust: Bayesian Prior-Guided Optimization \\ for Visual Generation}

\author{Ruiying Liu $^{1,2}$\thanks{Equal contribution.} , Yuanzhi Liang $^{2*}$, Haibin Huang$^{2}$, Tianshu Yu$^{1}$\thanks{Corresponding to yutianshu@cuhk.edu.cn; zhangc120@chinatelecom.cn.}, Chi Zhang$^{2\dagger}$ \\
$^1$The Chinese University of Hong Kong, Shenzhen, China \\
$^2$Institute of Artificial Intelligence, China Telecom (TeleAI), China \\
}

\iclrfinalcopy 
\begin{document}

\maketitle

\begin{abstract}
Group Relative Policy Optimization (GRPO) has emerged as an effective and lightweight framework for post-training visual generative models. However, its performance is fundamentally limited by the ambiguity of textual–visual correspondence: a single prompt may validly describe diverse visual outputs, and a single image or video may support multiple equally correct interpretations. This many-to-many relationship leads reward models to generate uncertain and weakly discriminative signals, causing GRPO to underutilize reliable feedback and overfit noisy ones.
We introduce Bayesian Prior-Guided Optimization (BPGO), a novel extension of GRPO that explicitly models reward uncertainty through a semantic prior anchor. BPGO adaptively modulates optimization trust at two levels: inter-group Bayesian trust allocation emphasizes updates from groups consistent with the prior while down-weighting ambiguous ones, and intra-group prior-anchored renormalization sharpens sample distinctions by expanding confident deviations and compressing uncertain scores.
Across both image and video generation tasks, BPGO delivers consistently stronger semantic alignment, enhanced perceptual fidelity, and faster convergence than standard GRPO and recent variants.
\end{abstract}
\section{Introduction}
Recent progress in text-to-video/image generation has been largely driven by powerful diffusion architectures~\cite{rombach2022high, ramesh2022hierarchical, saharia2022photorealistic, zhang2024vast} and reinforcement learning (RL)–based post-training strategies~\cite{black2023training,fan2023dpok,xue2025dancegrpo,liang2025integrating} that align generative models with perceptual or preference feedback~\cite{ouyang2022training,xu2024visionreward}. Among these, Group Relative Policy Optimization (GRPO)~\cite{shao2024deepseekmath,guo2025deepseekcoder}  has emerged as a promising framework, providing stable optimization and noticeable gains in visual quality, motion smoothness, and temporal coherence~\cite{xue2025dancegrpo}. However, despite these advances, the semantic alignment between text prompts and generated videos and images remains a persistent weakness—models often produce visually plausible yet semantically mismatched results.

\begin{figure}[tb]
\centering
    \begin{subfigure}[b]{.45\linewidth}
    \centering
    \includegraphics[width=.7\linewidth]{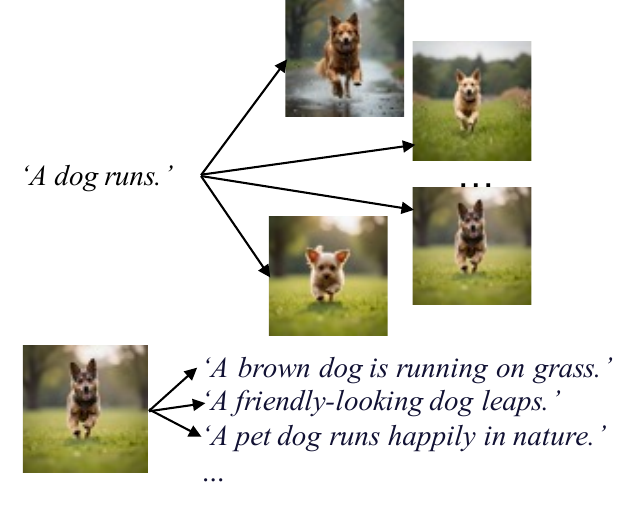}
    \caption{Textual-visual alignment is inherently many-to-many: a single generated image or video admits multiple semantically equivalent yet lexically diverse textual descriptions, a nuance often overlooked by conventional evaluation metrics.}
    \label{fig:TA_problem}
    \end{subfigure}
    \begin{subfigure}[b]{.45\linewidth}
    \centering
    \includegraphics[width=\linewidth]{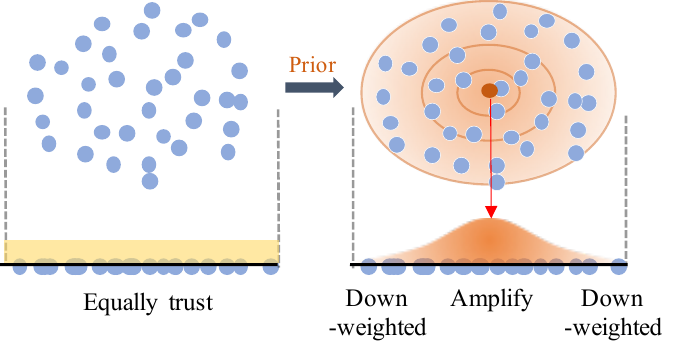}
    \caption{GRPO treats samples and rewards uniformly, the high uncertainty of alignment metrics intensify the underuse more useful rewards and hyperfocus possible noises. Posing prior helps to pay more attention on better signals anchored by the prior.}
    \label{fig:prior}
    \end{subfigure}
    \caption{Introducing priors helps to address high uncertainty of textual-visual alignment.}
\end{figure}

This limitation stems from the inherently ambiguous nature of textual-visual correspondence. A single video can be described in multiple semantically valid ways depending on temporal granularity or linguistic focus. For instance, a short sequence showing a gymnast performing rotations could be described as “doing a gymnastics spin,” “performing a turn,” or “completing two rounds of rotation.” Conversely, a single text prompt may correspond to diverse videos and images that differ in motion trajectory, style, or camera framing, yet still satisfy the same description. These one-to-many and many-to-one relationships, as shown in Figure~\ref{fig:TA_problem}, make the alignment task inherently uncertain, leading the reward model to produce unreliable or noisy signals. Existing GRPO-based methods generally treat rewards as consistent scalar feedback, without accounting for such uncertainty in the textual-visual relation.

To address this challenge, we propose Bayesian Prior-Guided Optimization (BPGO), a principled framework that explicitly models reward uncertainty within a Bayesian formulation. Rather than equally trusting all prompt groups, BPGO dynamically reallocates learning trust based on the consistency between observed rewards and a semantic prior anchor that represents the model’s expected performance as shown in Figure~\ref{fig:prior}. Groups achieving above-prior rewards receive amplified update gains, while unreliable groups are softly down-weighted. Within each group, a reward renormalization mechanism further sharpens discriminability by stretching confident deviations around the prior and compressing ambiguous ones. Together, these mechanisms create a hierarchical, quality-aware optimization landscape that enhances both reward reliability and semantic alignment during GRPO training.

In summary, our contributions are threefold:
\begin{itemize}
    \item We identify the intrinsic ambiguity of textual-visual alignment, and introduce priors to reduce the uncertainty for better post-training.
    \item We propose BPGO, a Bayesian prior-guided optimization framework that adaptively reweights and renormalizes reward signals according to their uncertainty, introducing both group-level trust allocation and sample-level discriminability enhancement.

    \item We demonstrate that BPGO significantly improves textual-visual alignment and overall generation quality across multiple benchmarks, outperforming recent GRPO-based approaches including DanceGRPO while maintaining computational efficiency.
\end{itemize}
\section{Related Works}

\noindent \textbf{GRPO and Variants}
Group Relative Policy Optimization (GRPO) was introduced by DeepSeekMath as a memory-efficient alternative to Proximal Policy Optimization (PPO)~\cite{shao2024deepseekmath}. Unlike PPO~\cite{schulman2017proximal}, which requires a separate value network to estimate advantages, GRPO generates multiple responses per prompt and uses normalized group rewards as baselines, achieving approximately 50\% reduction in memory usage. This design aligns naturally with the comparative nature of reward models typically trained on pairwise preference datasets~\cite{ouyang2022training}. 

DeepSeek-R1 demonstrated GRPO's effectiveness when combined iteratively with supervised fine-tuning for developing advanced reasoning capabilities~\cite{guo2025deepseekcoder}. The algorithm's group-based advantage estimation provides significantly lower variance compared to traditional policy gradient methods like REINFORCE~\cite{williams1992simple}, while avoiding the computational complexity of trust region methods~\cite{schulman2015trust}. Beyond mathematical reasoning, GRPO has been successfully applied to various language tasks including instruction following~\cite{ouyang2022training} and text summarization~\cite{stiennon2020learning}, demonstrating its versatility as a general-purpose alignment technique. Recent theoretical analyses have examined GRPO's convergence properties under different feedback models, establishing favorable sample complexity bounds that highlight its efficiency advantages in large-scale model training~\cite{casper2023open}.

\noindent \textbf{Visual Generation with Reinforcement Learning}
Early RL approaches for visual generation established important foundations for policy-based optimization of diffusion models. Denoising Diffusion Policy Optimization (DDPO)~\cite{black2023training} reformulated the multi-step diffusion sampling process as a Markov Decision Process, enabling the application of policy gradient algorithms to optimize text-to-image models for downstream objectives such as aesthetic quality and compressibility. DDPO demonstrated that treating each denoising step as an action allows for more effective optimization compared to reward-weighted likelihood approaches. Concurrently, Diffusion Policy Optimization with KL regularization (DPOK)~\cite{fan2023dpok} proposed online reinforcement learning with explicit KL divergence constraints to prevent reward overfitting. However, both methods faced critical scalability challenges when applied to large and diverse prompt sets, limiting their practical deployment.

DanceGRPO~\cite{xue2025dancegrpo} pioneered the adaptation of GRPO to visual generation, establishing the first unified framework capable of seamless operation across multiple generative paradigms (diffusion models and rectified flows), tasks (text-to-image, text-to-video, image-to-video), and foundation models. By reformulating the Stochastic Differential Equations underlying visual diffusion processes and carefully selecting optimized timesteps and noise initialization strategies, DanceGRPO achieved up to 181\% improvements over baselines on established benchmarks including HPS-v2.1~\cite{wu2023human} and VideoAlign. For autoregressive image generation, STAGE~\cite{yuan2025stage} addresses the unique challenges of discrete visual tokens by mitigating contradictory gradients and unstable entropy during GRPO training, recognizing that visual tokens represent continuous semantics requiring specialized handling.

Foundation models have established the architectural basis for modern visual generation. Stable Diffusion~\cite{rombach2022high} pioneered latent diffusion by performing the diffusion process in compressed latent space rather than pixel space, significantly improving computational efficiency. DALL-E 2~\cite{ramesh2022hierarchical} introduced conditioning on CLIP embeddings through a learned prior, while Imagen~\cite{saharia2022photorealistic} demonstrated the effectiveness of large pre-trained language models as text encoders. For video generation, comprehensive surveys~\cite{melnik2024video,wang2025survey} document architectural innovations including Space-Time U-Net architectures and Diffusion Transformer (DiT)~\cite{peebles2023scalable} designs that operate on spacetime patches for improved temporal coherence.

Reward modeling has emerged as a critical component enabling effective RLHF in visual domains. Recent work has developed specialized reward models that account for modern high-resolution, longer-duration videos, addressing limitations of earlier preference datasets~\cite{liu2025improving}. VisionReward~\cite{xu2024visionreward} and HPS-v2.1~\cite{wu2023human} provide multi-dimensional assessments of visual quality, text-image alignment, and motion characteristics. These advances in GRPO-based visual generation combine computational efficiency with sophisticated modeling capabilities, enabling more robust alignment of generative models with human preferences while maintaining training scalability.
\section{Bayesian Prior-Guided Policy Optimization}
Our method is based on GRPO. GRPO has emerged as a promising post-training approaches. Formally, based on recent research showing that the KL term can be neglected~\cite{xue2025dancegrpo}, the GRPO loss is formulated as 
\begin{equation}
    \mathcal{L}_\text{GRPO} = -\frac{1}{G} \sum_{i=1}^G \frac{1}{T}\sum_{t=0}^{T-1}h(r, \hat{A}(R), \theta, \epsilon)\label{eq:base_loss}
\end{equation}
where
\begin{equation*}
    h(r, \hat{A}, \theta, \epsilon) =\min \left(r_t^i(\theta) \hat{A}_t^i, \operatorname{clip}\left(r_t^i(\theta), 1-\epsilon, 1+\epsilon\right) \hat{A}_t^i\right),
\end{equation*}
\begin{equation*}
    \hat{A}_t^i=\frac{r_i-\operatorname{mean}\left(\left\{R_i\right\}_{i=1}^G\right)}{\operatorname{std}\left(\left\{R_i\right\}_{i=1}^G\right)}.
\end{equation*}
We denotes loss for the $i$-th prompt as $\mathcal{L}_i$, and it is calculated with the base loss Equation~\ref{eq:base_loss}.

While GRPO effectively improves visual quality and temporal coherence through preference-based feedback~\cite{xue2025dancegrpo}, it assumes that all reward signals are equally reliable. In practice, however, different prompts and their generated samples vary significantly in difficulty, reward confidence, and textual consistency. Thus, we propose Bayesian Prior-Guided Policy Optimization (BPGO), a modified version of GRPO designed to handle reward uncertainty and semantic ambiguity in text-to-video generation. 
BPGO introduces two complementary modules---\emph{Reliability-Adaptive Scaling (RAS)} and \emph{Contrastive Reward Transformation (CRT)}---that jointly improve reward reliability and optimization stability. RAS reallocates learning strength according to group-level reward reliability, while CRT enhances intra-group discriminability by stretching rewards relative to a semantic prior baseline. In all, these modules form a Bayesian-consistent optimization process that learns what to trust and how strongly to update during reinforcement fine-tuning.

\subsection{Semantic Prior and Adjustment}
We define a \emph{semantic prior} $R_{\text{prior}}$ to each sample, representing the expected reward of semantically clear and typical prompts. It can be estimated using calibration data or moving averages during training, and we introduce different prior rewards for different tasks, demonstrated in Section \ref{sec:experiment-setting}. For each reward $R$, the deviation $\Delta = R - R_{\text{prior}}$ quantifies how the observed quality diverges from prior expectation. A positive deviation indicates a reliable, well-aligned group, whereas a negative deviation suggests ambiguity or reward-model uncertainty. This prior-referenced deviation serves as an indicator of how much confidence the policy should place in each group’s gradients.

Our method performs two complementary adjustments under a unified Bayesian framework. The first adjustment centers rewards based on $\Delta_i = \bar{R}_{\text{group}} - R_{\text{prior}}$, where the prior reward $R_{\text{prior}}$ encodes our belief about expected output performance, while the observed group mean $\bar{R}_{\text{group}}$ provides empirical evidence from the current policy. This operation can correct for systematic bias by regularizing group statistics toward the prior. This corresponds to posterior mean adjustment in Bayesian inference. The second adjustment scales individual rewards relative to the prior. Specifically, this adjust each reward value in a group based on $\Delta_j = R^{(j)}_{\text{group}} - R_{\text{prior}}$ for $j=1, ... ,G.$, amplifying deviations to sharpen distinctions between outputs. This corresponds to precision modulation where rewards distant from the prior receive enhanced differentiation. These operations implement hierarchical Bayesian adjustment that corrects location bias at the group level while refining discrimination at the individual level. This dual mechanism reduces uncertainty in group statistics while enhancing signal strength for policy gradients, paralleling empirical Bayes methods that combine shrinkage with precision weighting.

\begin{figure}[tb]
    \centering
    \includegraphics[width=.98\linewidth]{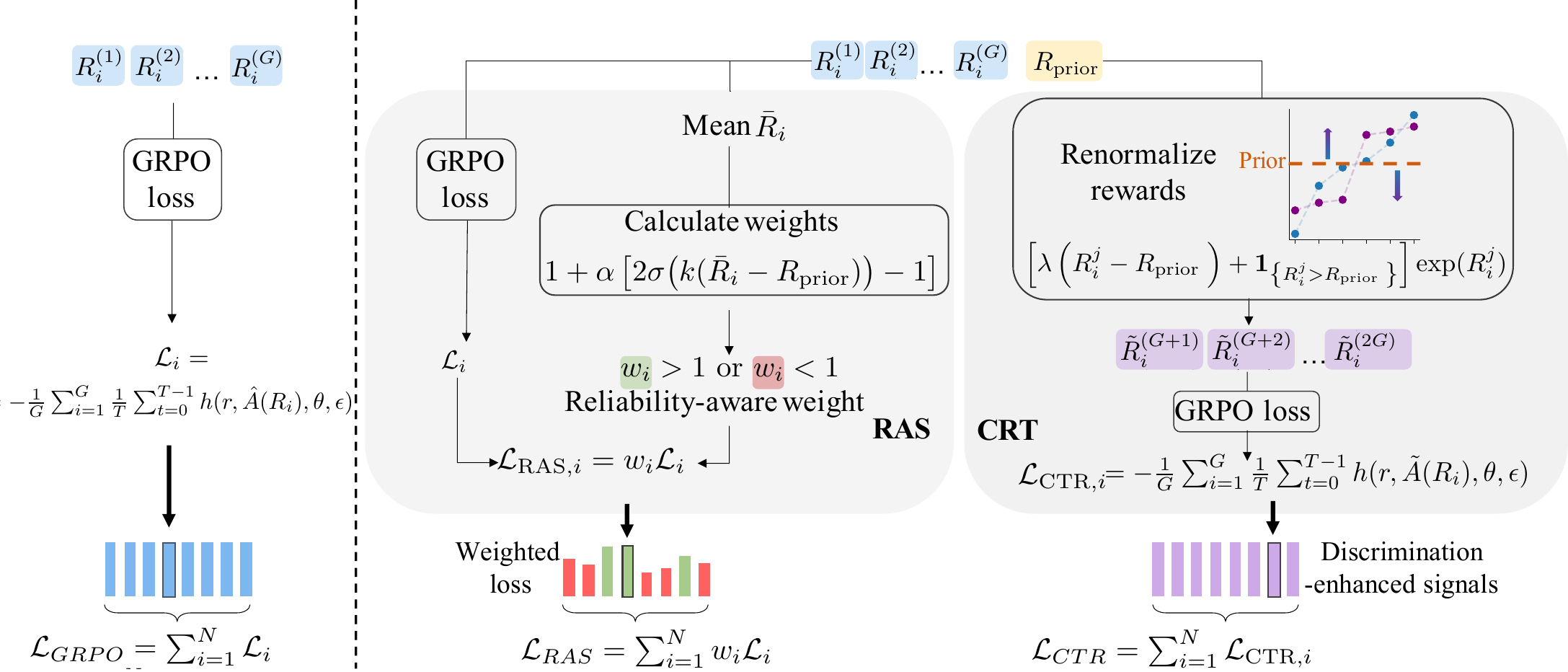}
    \caption{Enhanced GRPO framework incorporating Reliability-Adaptive Scaling (RAS) and Contrastive Reward Transformation (CRT). The architecture features two parallel branches: the RAS branch adaptively reweights samples based on their deviation from prior reward estimates, while the CRT branch generates an auxiliary reward group through reward renormalization to sharpen policy discrimination.}
    \label{fig:framework}
\end{figure}

\subsection{Reliability-Adaptive Scaling (RAS)}
Specifically, instead of a discrete piecewise rule, we adopt a smooth, differentiable trust function:
\begin{equation}
\begin{aligned}
        w_i &= f(\bar{R}_i - R_{\text{prior}})\\
        &=1 + \alpha \left[2\sigma\big(k(\bar{R}_i - R_{\text{prior}})\big) - 1\right],
\end{aligned}
\label{eq:RAS}
\end{equation}
where $\sigma(\cdot)$ denotes the sigmoid function, $\lambda$ controls the scaling magnitude, and $k$ adjusts the sharpness of transition. This continuous formulation behaves as a Bayesian update gain, dynamically modulating how much the posterior policy should rely on the current group’s evidence. When $\bar{R}_i > R_{\text{prior}}$, $w_{i} > 1$, amplifying gradients from high-confidence groups; when $\bar{R}_i < R_{\text{prior}}$, $w_{i} < 1$, softly reducing the impact of uncertain groups. Such reliability-adaptive weighting parallels trust-region ideas in policy optimization and uncertainty-aware learning strategies widely explored in both computer vision and reinforcement learning. The reweighted loss for group $i$ is defined as:
\begin{equation}
\mathcal{L}_{\text{RAS}, i} = w_{\text{group}, i} \cdot \mathcal{L}_{\text{GRPO}, i}.
\end{equation}
This mechanism directs the policy to focus its learning capacity on trustworthy semantic regions while avoiding overfitting to noisy or ambiguous prompts.

While GRPO is unaware beyond each group, which means it assumes that all samples are equally reliable. In practice, however, different prompts and their generated samples vary significantly in difficulty, reward confidence, and textual consistency. This mechanism implements a form of curriculum learning~\cite{bengio2009curriculum} where the model receives amplified gradients from ``mastered'' groups (high-quality directions worth reinforcing) and attenuated gradients from ``struggling'' groups (potentially noisy or suboptimal regions). 

The design also connects to recent findings in preference learning. Reweighting samples is shown effective in GRPO training for LLM training~\cite{zhou2025daro}, revealing that groups differ in their inherent difficulty, reward model reliability, and the current policy’s proficiency on different semantic categories. Thus, we distinguish groups containing more informative gradients and groups may introduce optimization noise based on the prior. Those groups with consistently higher rewards relative to baseline represent clearer preference signals, similar to how contrastive methods benefit from maximally separated pairs~\cite{yang2023rlcd,zhang2024videodpo}. By emphasizing these groups, we effectively increase sample efficiency---extracting more learning value from high-confidence training signals while being conservative with ambiguous ones.

\begin{algorithm}[tb]
\caption{BRPO}
\label{alg:brpo}
\begin{algorithmic}[1]
\Statex \textbf{Initialize:} Initial policy parameters $\theta_0$, group size $G$, reward function $r(o)$, and hyper-parameters.
\State $\theta \gets \theta_0$.
\For{each training step $t = 1, 2, \dots, T$}
    \For{each prompt $\text{prompt}_i$ in current batch $B$}
        \State Sample $G$ candidate outputs for each prompt $o_i=\{\text{o}_i^{(1)}, ...\text{o}_i^{(G)}\}$
        \State Rewards $R_i=\{R_i^{(1)}, ...R_i^{(G)}\} \gets r(o_i)$
        \State  $\mathcal{L}_{i} \gets R_i$ with Equation~\ref{eq:base_loss}
        \State \textcolor{blue}{Calculate $w_i$ with Equation~\ref{eq:RAS} // RAS}
        \State \textcolor{purple}{$\tilde{R}_i=\{\tilde{R}_i^{(1)}, ..., \tilde{R}_i^{(G)}\} \gets R_i$ with Equation~\ref{eq:CRT}}
        \State \textcolor{purple}{$\mathcal{L}_{\text{CRT},i} \gets \tilde{R}_i$ with Equation~\ref{eq:base_loss} // CRT}
    \EndFor
    \State \textcolor{blue}{$\mathcal{L}_{\text{RAS}} \gets \sum_{i=1}^{|B|}w_i\mathcal{L}_{\text{RAS},i}$// RAS}
    \State \textcolor{purple}{$\mathcal{L}_{\text{CRT}} \gets \sum_{i=1}^{|B|}\mathcal{L}_{\text{CRT},i}$ // CRT}
    \State $\mathcal{L} = \mathcal{L}_{\text{RAS}} +\beta\mathcal{L}_{\text{CRT}}$
    \State Calculate $\nabla_\theta \mathcal{L}_{\text{BRPO}}$, update $\theta$
\EndFor
\end{algorithmic}
\end{algorithm}
\subsection{Contrastive Reward Transformation (CRT)}
Within each prompt group, the raw reward distribution often exhibits small numerical differences, but the standardization operation in advantages calculation poses relative differences under the hypothesis of a Gaussion distribution, failing to reflect accurate signals. Also, this continuous normalization may not fully exploit the quality structure present in the data. 

We therefore introduce \emph{Contrastive Reward Transformation}, which expands reward deviations around the prior to introduce better distinctions:
\begin{equation}
\begin{aligned}
    \tilde{R}_i^{(j)} &=g(R_i^{(j)} - R_{\text{prior}}) \\
&= \left[\lambda\left(R_i^{(j)}-R_{\text {prior }}\right)+\mathbf{1}_{\left\{R_i^{(j)}>R_{\text {prior }}\right\}}\right] \operatorname{exp}(R_i^{(j)}),
\end{aligned}
\label{eq:CRT}
\end{equation}
where $\lambda > 0$ is a contrast factor, and $\lambda$. This transformation geometrically stretches the reward space, emphasizing samples that deviate confidently from the prior while compressing ambiguous ones. We compute an auxiliary loss based on $\tilde{R}_i^{(j)}$:
\begin{equation}
\mathcal{L}_{\text{CRT}, i} = \mathcal{L}_{\text{GRPO}}(\{\tilde{R}_i^{(j)}\}_{j=1}^G),
\end{equation}
which is added to the original loss.

\subsection{Overall Objective}
Combining RAS and CRT, the full training objective over $N$ prompt groups is:
\begin{equation}
\mathcal{L}_{\text{BPGO}} = \frac{1}{N}\sum_{i=1}^N \left[\mathcal{L}_{\text{RAS}, i} + \beta \mathcal{L}_{\text{CRT}, i}\right] 
\end{equation}
where $\beta$ is This hierarchical formulation realizes a Bayesian prior-guided policy update, where 
\emph{RAS} adjusts posterior trust across groups (macro-level reliability) and 
\emph{CRT} refines posterior sharpness within groups (micro-level contrast). 
The whole algorithm is shown in Algorithm~\ref{alg:brpo}.
\section{Experiment}
\subsection{Settings} \label{sec:experiment-setting}
\paragraph{Models} 
We evaluate our BPGO framework across three distinct visual generation tasks to demonstrate its generalizability. For text-to-video generation (T2V), we employ Wan2.1-1.3B, a compact yet efficient diffusion-based video generation model. For image-to-video generation (I2V), we utilize Wan2.2-14B, a significantly larger model incorporating mixture-of-experts (MoE) architecture with two 14B parameter modules, enabling high-fidelity video synthesis conditioned on initial frames. For text-to-image generation (T2I), we adopt FLUX, a state-of-the-art latent diffusion model known for its superior prompt adherence and visual quality. All models are initialized from their respective supervised fine-tuned (SFT) checkpoints to ensure strong baseline performance.

\paragraph{Priors} 
For text-to-video generation, we use rewards from the SFT model's generations as the prior, providing stable reference anchoring for optimization. For image-to-video generation with large models (e.g., Wan2.2 14B), we leverage the first frame's text-alignment as a natural baseline, where any generated video should maintain alignment quality present in the conditioning frame. For text-to-image generation, we employ the running mean of group rewards as the prior, which accumulates historical information to capture evolving quality distributions while smoothing transient fluctuations.

\paragraph{Hyper-parameters} 
We set the group size $G=8$ for video generation and $G=12$ for image generation. For video generation we adopt VideoCLIP-XL~\cite{wang2024videoclip} as reward models, and we use HPSv2~\cite{
wu2023human} as rewards for image generation. More settings can be found in Appendix~\ref{supp:addtional-setting}.

\begin{wraptable}{R}{0.6\textwidth}
    \begin{minipage}{0.6\textwidth}
            \centering
            \caption{Results on differrent tasks. \textdagger Our implemented DanceGRPO.}
            \tabcolsep=0.08cm 
            \begin{tabular}{cc|ccc}
            \toprule
                Task & Method & VideoClipXL  & \makecell{VideoAlign\\-TA} & \makecell{VideoAlign\\-overall}\\
                \midrule
                \multirow{3}{*}{T2V}
                & Wan2.1 & 2.6563 & 1.0638 & 0.0939 \\
                & GRPO\textdagger & 2.6714 & 0.8984 & -0.5411\\
                & Ours & 2.6788 & 1.1193 & -0.0478\\
                \midrule
                \multirow{3}{*}{I2V}
                & Wan2.2 & 2.6726 & 1.0633 & -0.7623 \\
                & GRPO\textdagger & 2.0713 & 0.2307 & -1.8932 \\
                & Ours & 2.6855 & 1.0589 & -1.0491\\
                \midrule
                \midrule
                & & \makecell{HPSv2} & PickScore & ImageReward\\
                \midrule
                \multirow{3}{*}{T2I} 
                & FLUX & 0.2398 & 0.2270 & 1.1482 \\
                & GRPO\textdagger & 0.2564 & 0.2242 & 1.0607\\
                & Ours & 0.2434 & 0.2288 & 1.2136\\
            \bottomrule
            \end{tabular}
            \label{tab:sota}
    \end{minipage}
    \vspace{-11pt}
\end{wraptable}
\subsection{Results}
Table~\ref{tab:sota} reports quantitative results on text-to-video (T2V), image-to-video (I2V), and text-to-image (T2I) generation. Across all settings, BPGO achieves consistent improvements over both the pretrained base models and the GRPO post-training baseline. The gains are particularly pronounced on alignment-focused metrics (VideoAlign-TA~\cite{liu2025improving}, PickScore~\cite{kirstain2023pick}, ImageReward~\cite{xu2023imagereward}), which are known to be highly sensitive to semantic mismatches and reward noise. This suggests that BPGO effectively enhances reward discriminability in scenarios where textual–visual correspondence is uncertain.

\noindent \textbf{Text-to-Video (T2V).}
BPGO delivers an improvement on VideoCLIP-XL from 2.6714 (GRPO) to 2.6788, indicating a stronger global alignment between the generated video and the textual prompt. More importantly, VideoAlign-TA increases from 0.8984 to 1.1193 (+24.6\%), reflecting significantly sharper, more reliable textual–visual alignment. VideoAlign-overall also improves substantially from $-0.5411$ to $-0.0478$, reducing the penalty associated with ambiguous reward predictions. These results suggest that BPGO mitigates both the under-discriminative reward issue and the instability caused by noisy alignment metrics.

\noindent \textbf{Image-to-Video (I2V).}
In I2V tasks, where the model must preserve the input image identity while generating coherent motion, BPGO improves VideoCLIP-XL from 2.6726 to 2.6855. Despite the stricter constraints in I2V generation—where reward noise is typically amplified due to complex motion and long temporal horizons—BPGO maintains a stable VideoAlign-TA (1.0589) and achieves competitive VideoAlign-overall performance. The results demonstrate that BPGO effectively removes ambiguous samples from dominating the learning dynamics, enabling more stable optimization across multiple modalities.

\noindent \textbf{Text-to-Image (T2I).}
For T2I generation, BPGO enhances both PickScore (0.2242 $\rightarrow$ 0.2288) and ImageReward (1.0607 $\rightarrow$ 1.2136). The improvement in ImageReward—a human preference–aligned metric—is especially noteworthy, as it highlights BPGO’s ability to amplify reliable reward signals and suppress subtle noise that often misleads continuous reward models. While the absolute gains on HPSv2 are more modest, this is expected given that T2I rewards tend to saturate at higher quality levels; nevertheless, the improved PickScore indicates better fine-grained object–attribute binding.

\begin{figure}[tb]
    \centering
    \includegraphics[width=\linewidth]{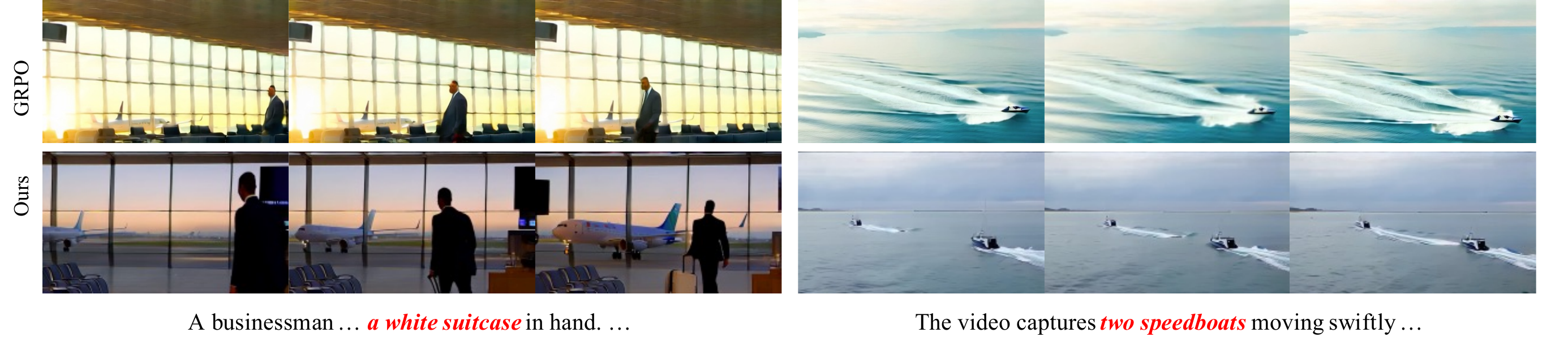}
    \caption{Visual comparison on text-to-video task. For each videos, the top row is GRPO, and the bottom row is BPGO.}
    \label{fig:vis_t2v}
\end{figure}
\begin{figure}[tb]
    \centering
    \includegraphics[width=\linewidth]{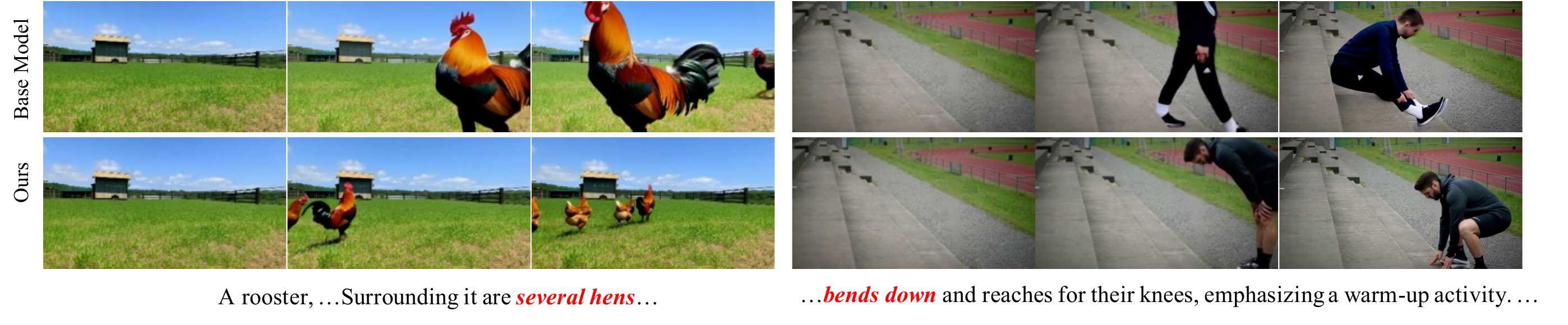}
    \caption{Visual comparison on image-to-video task. For each videos, the top row is base model, and the bottom row is BPGO.}
    \label{fig:vis_i2v}
\end{figure}
\begin{figure}[tb]
    \centering
    \includegraphics[width=.9\linewidth]{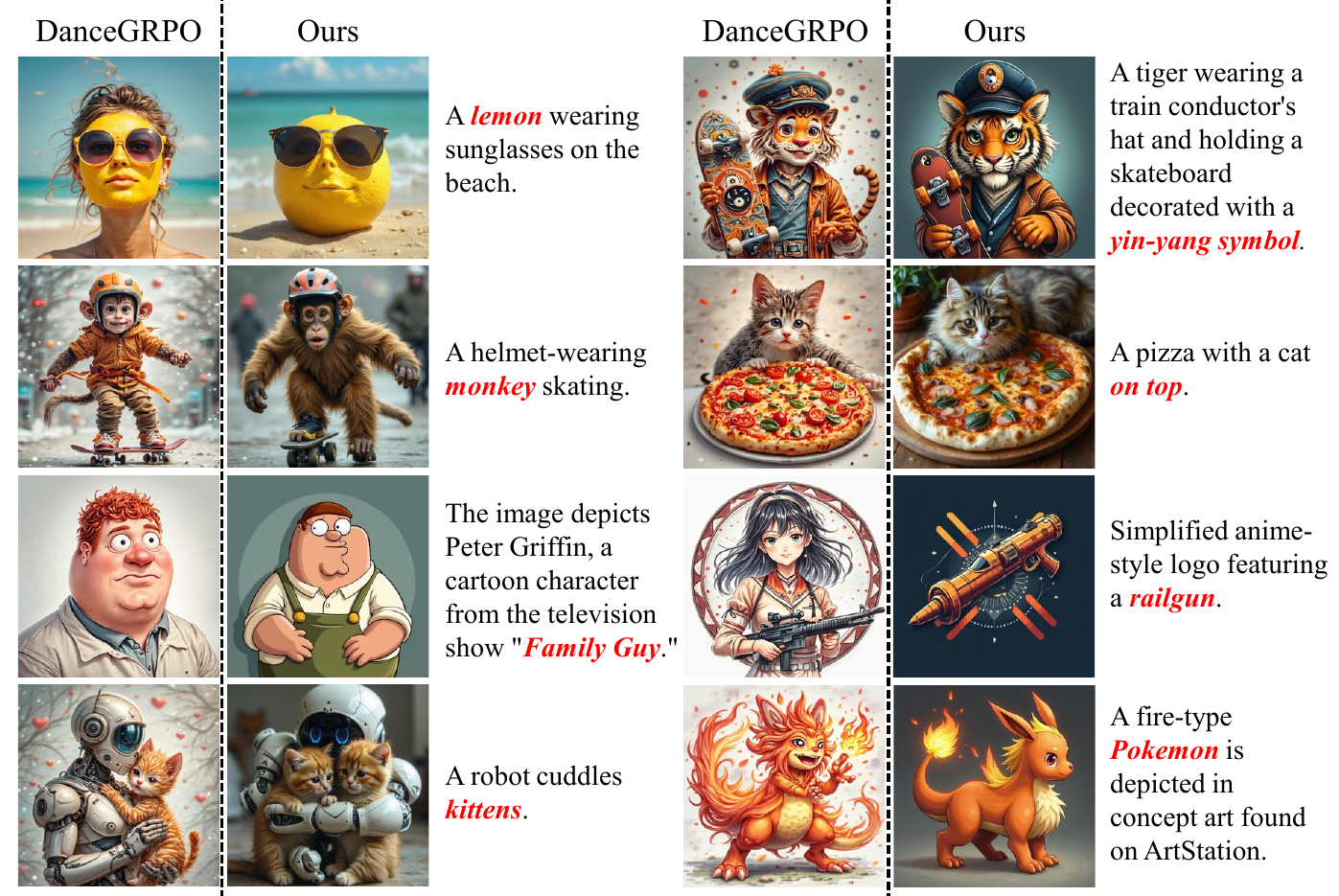}
    \caption{Visual comparison on text-to-image task. For each pair, the left is GRPO, and the right is BPGO.}
    \label{fig:vis_t2i}
\end{figure}
\noindent \textbf{Qualitative Comparison.}
Figures~\ref{fig:vis_t2v}, \ref{fig:vis_i2v} and \ref{fig:vis_t2i} provide qualitative evidence supporting the quantitative findings. In T2I, GRPO often generates semantically loose interpretations or misplaces fine-grained details, whereas BPGO consistently produces images with sharper object boundaries, more coherent spatial relations, and semantically accurate attributes. This aligns with observations that reward ambiguity in GRPO causes the model to underfit high-precision details, while BPGO’s prior-guided renormalization sharpens discriminative signals.

For I2V, BPGO preserves the appearance of the input image more faithfully and generates smoother, more purposeful motion. GRPO tends to produce temporal inconsistencies such as jitter, abrupt motion transitions, or partial identity drift. BPGO mitigates these artifacts by amplifying confident reward deviations during training, guiding the model toward consistently meaningful updates across frames.

Together, these numerical and visual results show that BPGO significantly improves alignment quality, visual fidelity, and structural coherence in both image and video generation. More results can be found in Appendix~\ref{supp:addtional-results}.

\subsection{Reward Dynamics and Ablation Studies}

\begin{figure}[tb]
    \centering
    \begin{minipage}{0.47\linewidth}
    \centering
        \includegraphics[width=\linewidth]{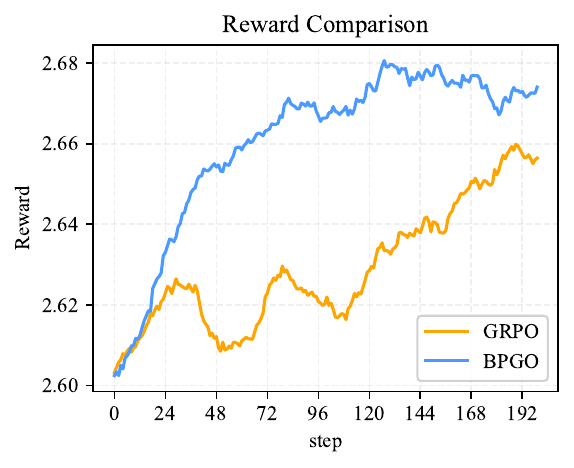}
        \caption{Reward curve comparison on text-to-video task.}
        \label{fig:reward-curve-t2v}
    \end{minipage}
    \begin{minipage}{0.47\linewidth}
        \centering
        \includegraphics[width=\linewidth]{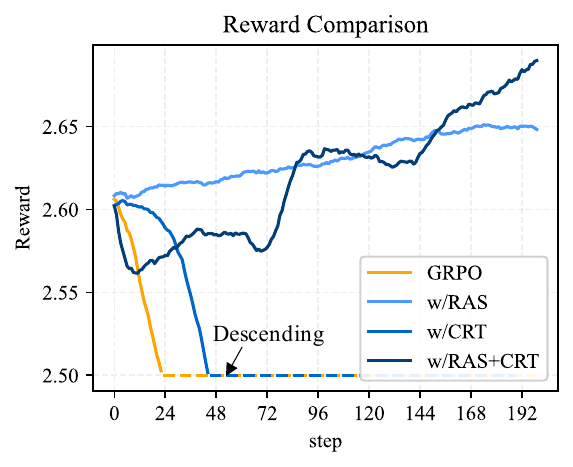}
        \caption{Reward curve comparison on image-to-video task.}
        \label{fig:reward-curve-i2v}
    \end{minipage}
\end{figure}

\noindent \textbf{Reward Optimization Behavior.}
Figure~\ref{fig:reward-curve-t2v} shows the reward progression during T2V training. GRPO exhibits slow convergence and noticeable oscillations, characteristic of optimization under noisy continuous rewards. In contrast, BPGO achieves (1) a steeper increase in early reward, indicating more effective credit assignment, (2) smoother reward trajectories due to reduced sensitivity to ambiguous samples, and (3) higher final reward, reflecting improved policy quality. This behavior matches prior observations in RL that Bayesian regularization stabilizes training under uncertain rewards.

\textbf{Ablation on RAS and CRT.}
Table~\ref{tab:ablation} examines the contributions of the Reward Adaptation Strategy (RAS) and the Confidence-based Reward Transformation (CRT). RAS alone improves VideoAlign-TA from baseline GRPO by +0.1491 (T2V) and +0.1675 (I2V), confirming the effectiveness of group-level trust modulation. CRT alone boosts VideoAlign-overall and enhances discriminability but can cause training instability, as its aggressive score stretching amplifies both informative and noisy deviations.\begin{wraptable}{R}{0.6\textwidth}    
    \vspace{-12pt}
    \begin{minipage}{0.6\textwidth}
        \centering
        \caption{Ablation study.}
        \tabcolsep=0.01cm 
        \begin{tabular}{cc|ccc}
        \toprule
            Task & Method & VideoClipXL  & \makecell{VideoAlign\\-TA} & \makecell{VideoAlign\\-overall}\\
            \midrule
            \multirow{3}{*}{T2V}
            & RAS & 2.7042 & 1.2327 & -0.4838\\
            & CRT & 2.6844 & 1.1751 & 0.0876\\
            & RAS+CRT & 2.6788 & 1.1193 & -0.0478\\
            \midrule
            \multirow{3}{*}{I2V}
            & RAS & 2.6681 & 1.0361 & -0.8429 \\
            & CRT & 2.0682 & 0.2162 & -1.6573 \\
            & RAS+CRT & 2.6855 & 1.0589 & -1.0491\\
            \midrule
            \midrule
            & & \makecell{HPSv2} & PickScore & ImageReward\\
            \midrule
            \multirow{3}{*}{T2I} 
            & RAS & 0.2578 &  0.2244 & 1.0788 \\
            & CRT & 0.2562 & 0.2273 & 1.2375 \\
            & RAS+CRT & 0.2434 & 0.2288 & 1.2136\\
        \bottomrule
        \end{tabular}
        \label{tab:ablation}
    \end{minipage}
\vspace{-15pt}
\end{wraptable}

The combined RAS+CRT achieves the best results in most metrics across modalities. This synergy arises because RAS provides global stability by adapting trust to prior consistency, while CRT sharpens local discriminability by expanding meaningful deviations. Figure~\ref{fig:reward-curve-i2v} further validates this: CRT alone accelerates early optimization but suffers severe oscillations, whereas RAS alone yields gradual but stable progress. Their combination balances these strengths, achieving both robustness and rapid convergence.

\noindent \textbf{Interpretation.}
The ablation trends support the central hypothesis behind BPGO: textual–visual alignment rewards are inherently noisy and ambiguous, requiring both global trust adjustment and local discriminability enhancement. RAS effectively suppresses ambiguous group-level patterns, while CRT focuses learning on confident deviations, creating a more informative reward landscape. The complementary nature of these components provides empirical evidence for the necessity of a hierarchical Bayesian optimization framework.

\subsection{Hyper-parameter Analysis and Discussion}
\textbf{Sensitivity to Prior Weighting $\alpha$.} Table~\ref{tab:alpha} analyzes the effect of the RAS scaling parameter $\alpha$ in Equation~\ref{eq:RAS}, which controls how strongly rewards are reweighted. The results reveal a clear performance peak at $\alpha=0.5$ on both VideoCLIP-XL and VideoAlign-TA. Smaller $\alpha$ yields insufficient emphasis on strong reward signals, while larger $\alpha$ over-amplifies noise, degrading stability. This trade-off resembles phenomena observed in uncertainty-aware RL, where moderate weighting leads to optimal generalization.

\begin{wraptable}{r}{0.45\textwidth}
    \begin{minipage}{0.45\textwidth}
    \centering
    \caption{Text-to-Video results on different $\alpha$.}
    \tabcolsep=0.02cm 
    \begin{tabular}{c|ccc}
    \toprule
        $\alpha$ & VideoClipXL & \makecell{VideoAlign\\-TA} & \makecell{VideoAlign\\-overall} \\
        \midrule
        0.1 & 2.6834 & 1.1866 & 0.0386\\
        0.5 & 2.7042 & 1.2327 & -0.4838\\
        0.7 & 2.6678 & 1.1161 & 0.0979\\
        0.9 & 2.6583 & 1.0334 & -0.0120\\
    \bottomrule
    \end{tabular}
    \label{tab:alpha}
    \end{minipage}
\end{wraptable} 
Across all tasks and metrics, three core principles of BPGO emerge:
(1) \emph{Reward uncertainty modeling is essential.} Alignment rewards are inherently noisy due to many-to-many textual–visual correspondence. BPGO explicitly accounts for this by anchoring optimization to a semantic prior.  
(2) \emph{Hierarchical modulation improves learning.} Group-level trust and sample-level discriminability act on different statistical structures in reward distributions, collectively producing a more structured optimization landscape.  
(3) \emph{Regularization through dual losses enhances robustness.} Combining original and renormalized rewards improves stability, prevents overfitting to noisy patterns, and encourages consistent updates across samples.

These observations align with findings in distributional RL and calibrated reward modeling: improving reward structure and reducing variance leads to more robust and interpretable policy updates. The consistent improvements of BPGO across image and video tasks demonstrate that prior-guided uncertainty modeling is a powerful strategy for post-training large-scale visual generative models.
\section{Conclusion}
In this work, we addressed a fundamental limitation of GRPO-based post-training for visual generative models: the inherent ambiguity of textual–visual correspondence and the resulting uncertainty in reward signals. We introduced Bayesian Prior-Guided Optimization (BPGO), a principled extension of GRPO that incorporates a semantic prior anchor to guide optimization under uncertain or weakly discriminative feedback. Through inter-group Bayesian trust allocation and intra-group prior-anchored renormalization, BPGO constructs a prior-aware optimization that emphasizes reliable signals while gracefully handling ambiguous or contradictory ones.
Extensive experiments on both image and video generation tasks demonstrate that BPGO improves textual–visual alignment and convergence behavior over GRPO and recent variants such as DanceGRPO, while preserving computational efficiency. Together, these results highlight the importance of uncertainty-aware optimization in reinforcement learning for generative models and position BPGO as a robust and scalable approach for aligning high-capacity visual generators with complex semantic objectives.

\bibliography{iclr2026_conference}
\bibliographystyle{iclr2026_conference}

\appendix
\clearpage
\setcounter{page}{1}
\section{Addtional Results}\label{supp:addtional-results}
\subsection{Text-to-Video results on VBench.}
\begin{table}[htbp]
    \centering
    \caption{Text-to-video results on VBench video-condition consistency metrics (Part 1). We compare our proposed modules against the baseline GRPO. The best results are shown in bold, and the second-best results are underlined. \textdagger Our implemented DanceGRPO.}
    \tabcolsep=0.03cm 
    \begin{tabular}{c|ccccccccc}
    \toprule
        Method & \makecell{Object\\Class} & \makecell{Multiple\\Objects} & \makecell{Human\\Action} & Color & Scene & \makecell{Temporal\\Style} & \makecell{Overall\\Consistency} & \makecell{Appearance\\Style} & \makecell{Spatial\\Relationship}\\
        \midrule
        GRPO\textdagger & \underline{0.6875} & 0.2111 & 0.6100 & \underline{0.8408} & 0.1628 & 0.2285 & 0.2179 & \underline{0.2000} & \textbf{0.3320} \\
        RAS & 0.4902 & 0.1021 & 0.5580 & 0.8127 & 0.1302 & \textbf{0.2386} & \textbf{0.2306} & \textbf{0.2066} & 0.2565\\
        CRT & 0.5434 & \underline{0.2485} & \underline{0.6300} & 0.8218 & \textbf{0.2278} & 0.2252 & 0.2264 & 0.1912 & 0.2900\\
        RAS+CRT & \textbf{0.6899} & \textbf{0.2736} & \textbf{0.6500} & \textbf{0.8594} & \underline{0.2253} & \underline{0.2342} & \underline{0.2301} & 0.1964 & \underline{0.2907}\\
    \bottomrule
    \end{tabular}
    \label{tab:vbench}
\end{table}
We evaluated the results using video-condition consistency metrics from VBench~\cite{huang2024vbench}, which encompasses 9 dimensions. The results are shown in Table~\ref{tab:vbench}. Our BPGO with the complete two modules (RAS+CTR) achieves the best performance across most dimensions, demonstrating significant improvements over the baseline GRPO. Specifically, our BPGO obtains the highest scores in four metrics. The individual modules also show competitive performance, with RAS excelling in Appearance Style and CTR achieving strong results in Scene. These results validate the effectiveness of our proposed modules, particularly when combined, in enhancing video-condition consistency across diverse evaluation criteria.

\subsection{Additional Qualitative Comparison.}
\begin{figure}
    \centering
    \includegraphics[width=.9\linewidth]{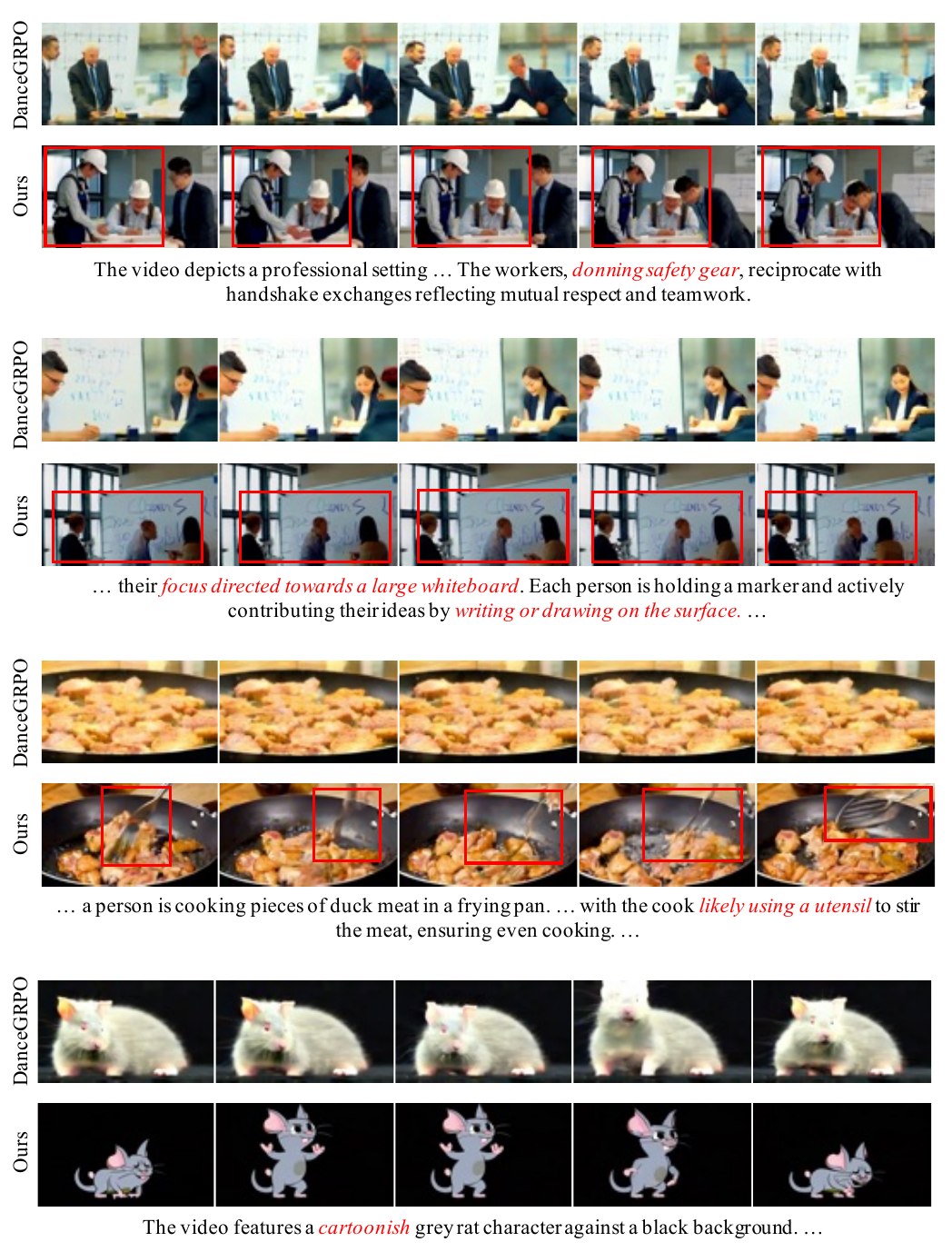}
    \caption{Qualitative comparison on text-to-video task.}
    \label{fig:app-vis_t2v1}
\end{figure}

\begin{figure}
    \centering
    \includegraphics[width=.9\linewidth]{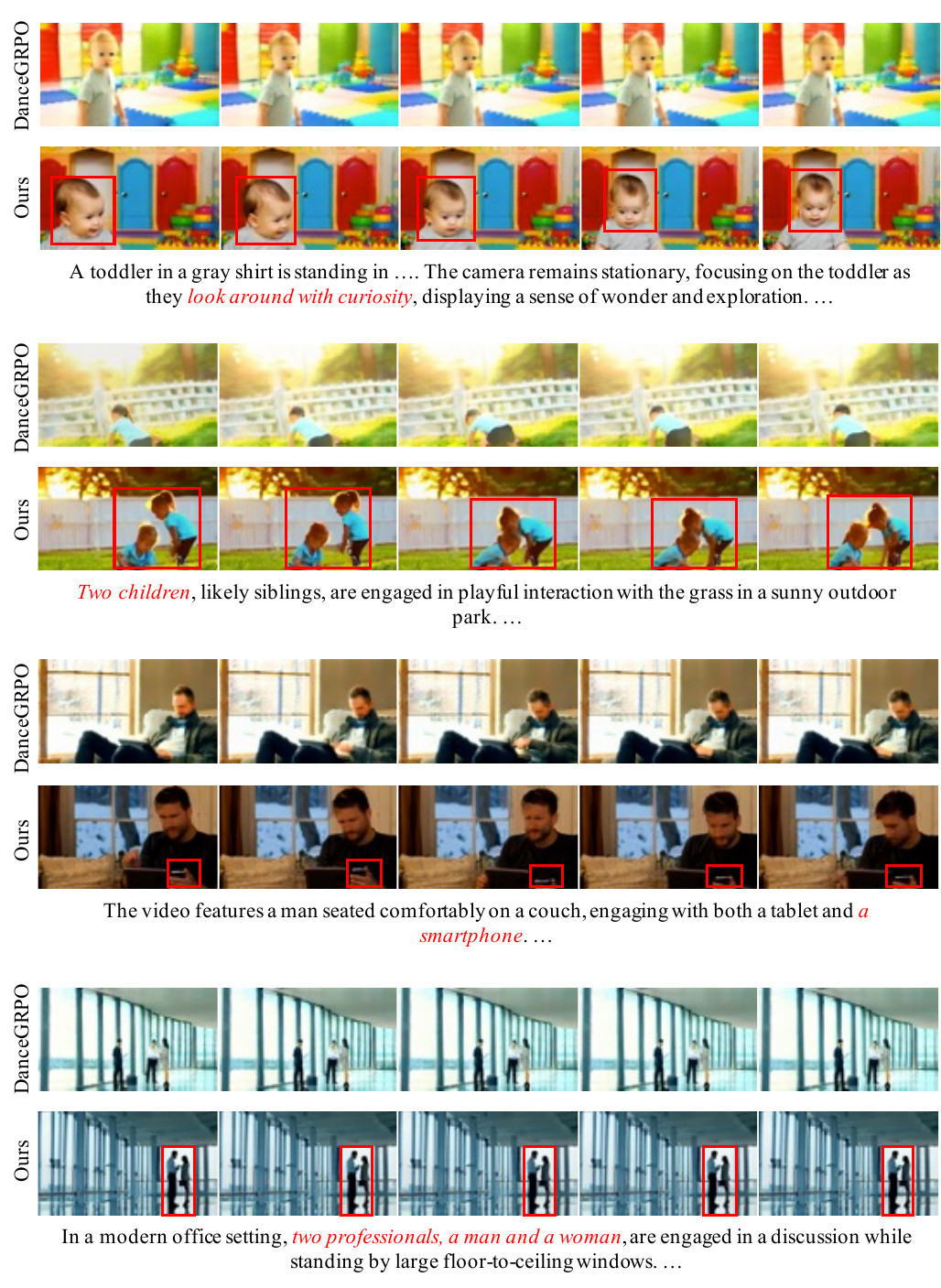}
    \caption{Qualitative comparison on text-to-video task.}
    \label{fig:app-vis_t2v2}
\end{figure}

\begin{figure}
    \centering
    \includegraphics[width=.9\linewidth]{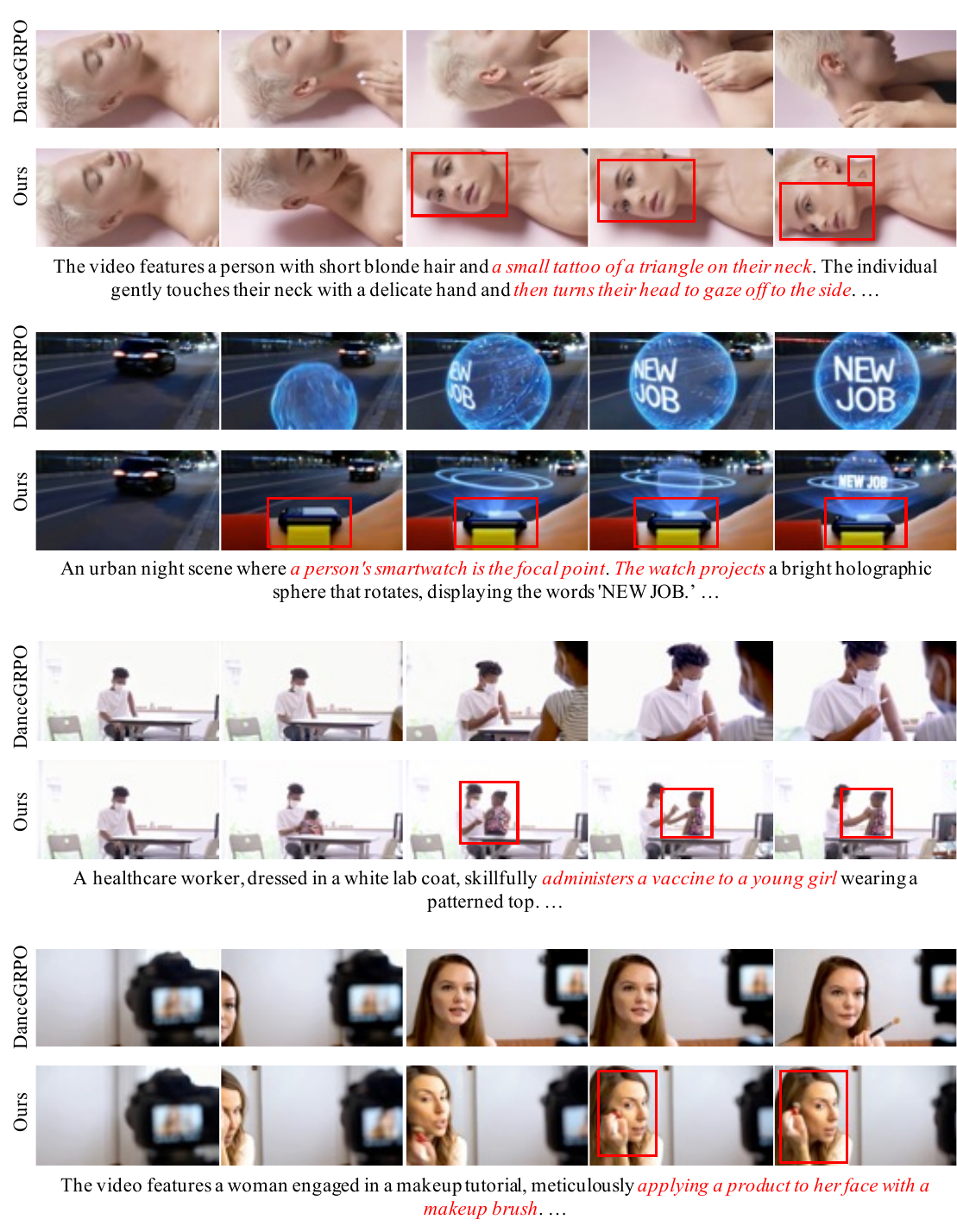}
    \caption{Qualitative comparison on image-to-video task.}
    \label{fig:app-vis_i2v1}
\end{figure}

\begin{figure}
    \centering
    \includegraphics[width=.9\linewidth]{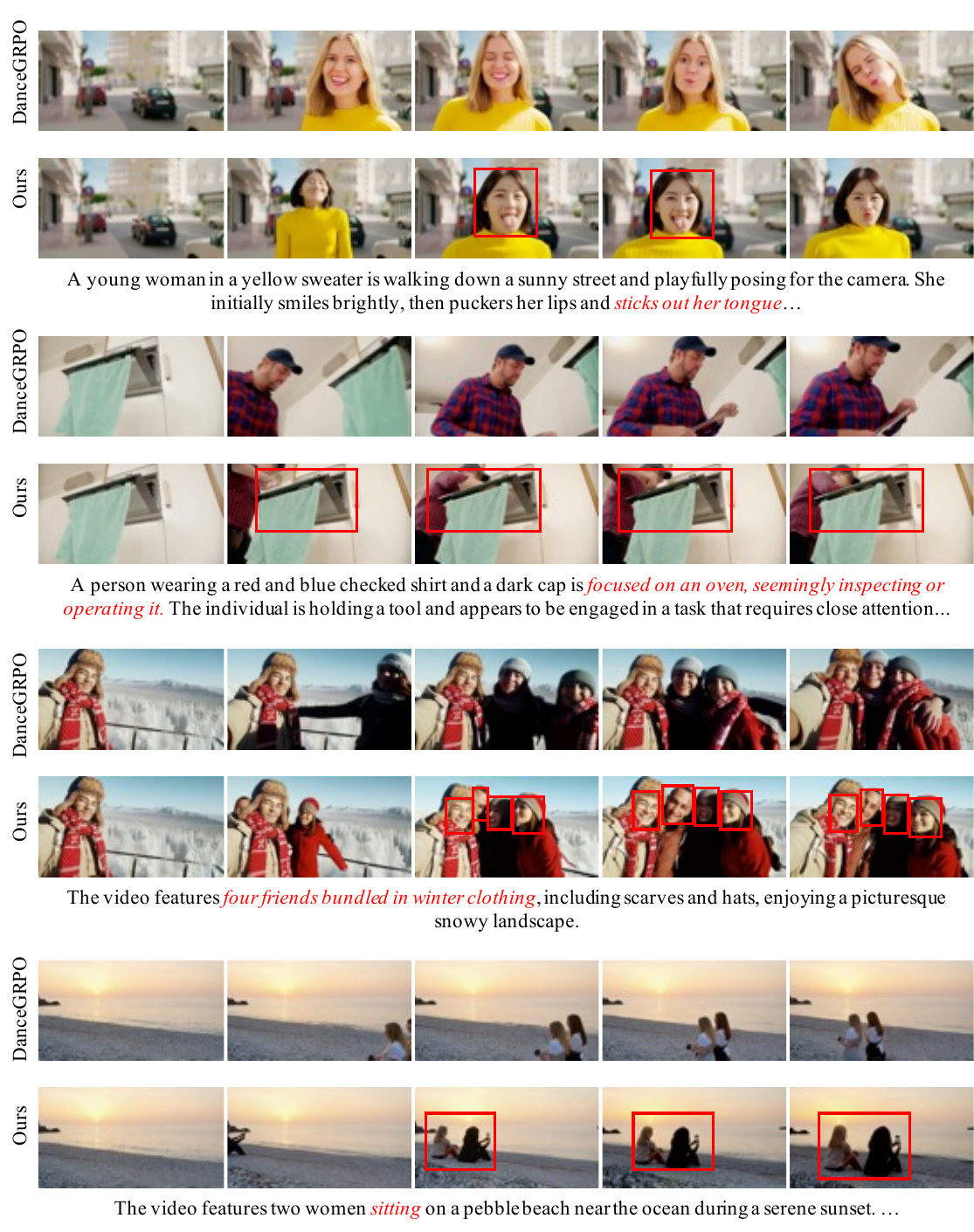}
    \caption{Qualitative comparison on image-to-video task.}
    \label{fig:app-vis_i2v2}
\end{figure}

Figure~\ref{fig:app-vis_t2v1}, \ref{fig:app-vis_t2v2}, \ref{fig:app-vis_i2v1} and \ref{fig:app-vis_i2v2} present qualitative comparisons on text-to-video and image-to-video generation tasks. Our method consistently outperforms DanceGRPO in maintaining consistency with input conditions. For text-to-video generation, our approach accurately captures critical details such as safety equipment, specific actions (e.g., writing, cooking with utensils), object attributes (e.g., cartoonish appearance), and correct object counts (e.g., two children). For image-to-video generation, our method preserves fine-grained visual details while correctly generating specified attributes (e.g., triangle tattoo, "NEW JOB" text projection), actions (e.g., vaccine administration, makeup application), and multi-subject scenes (e.g., four friends in winter clothing), as highlighted by the red bounding boxes. These examples illustrate our model's superior ability to generate videos that accurately reflect semantic content and fine-grained details from both textual and visual conditions.

\begin{figure}
    \centering
    \includegraphics[width=.9\linewidth]{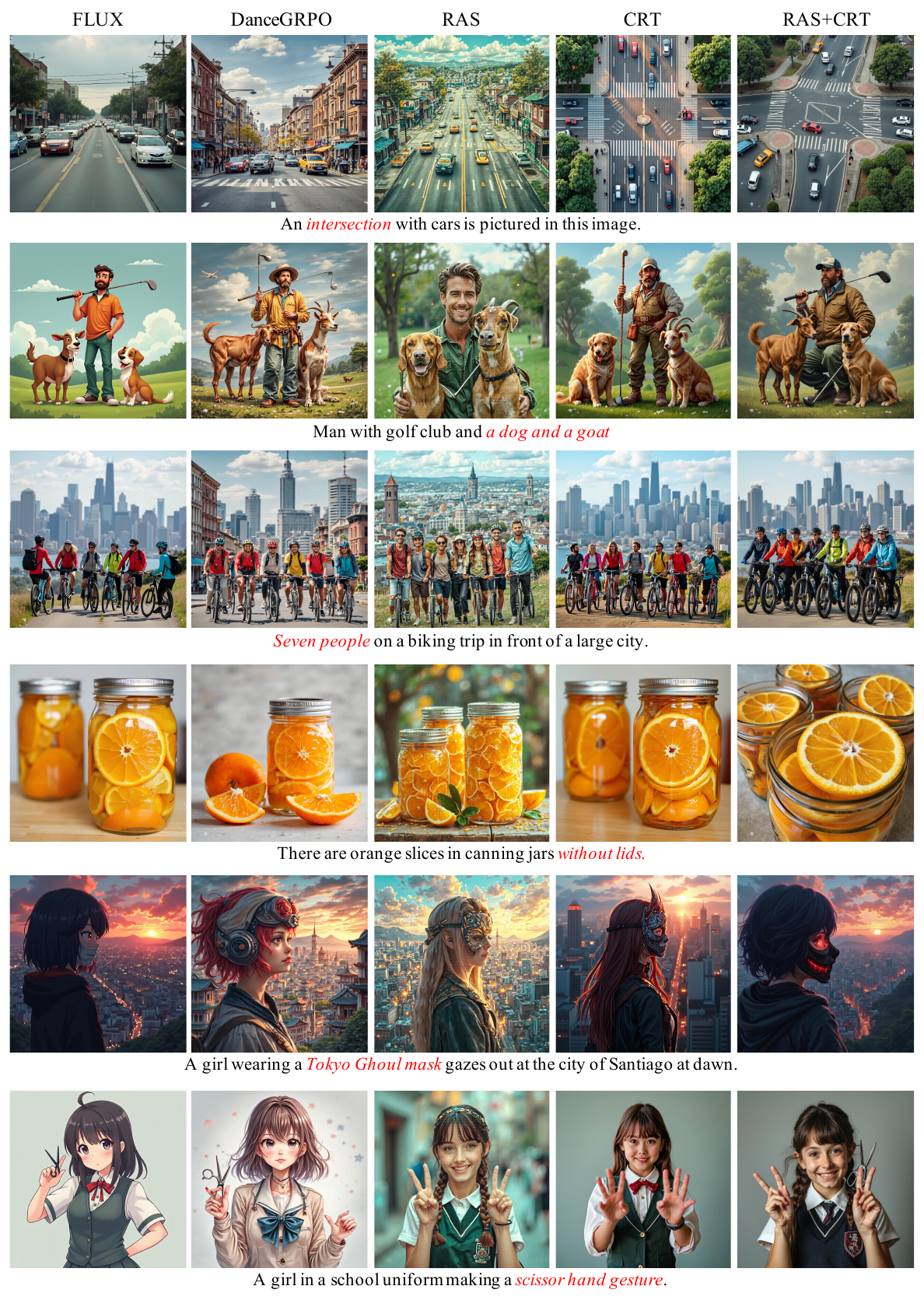}
    \caption{Qualitative comparison on text-to-image task.}
    \label{fig:app-vis_t2i}
\end{figure}

Figure~\ref{fig:app-vis_t2i} shows qualitative results on text-to-image generation. Our method (RAS+CRT) outperforms baselines in accurately reflecting textual details. Key improvements include: correct object counts (e.g., "a dog and a goat," "seven people"), fine-grained attributes (e.g., "Tokyo Ghoul mask," "without lids"), and specific gestures (e.g., "scissor hand gesture"). While FLUX and DanceGRPO often miss critical details or generate incorrect quantities, our combined approach consistently produces images that faithfully align with the text prompts.

\section{Additional Experiment Settings}\label{supp:addtional-setting}
\subsection{Hyperparameter Settings}
We conduct experiments across three generation tasks with the following configurations. For text-to-video generation, we use a batch size of 32 distributed across 32 H100 GPUs. For image-to-video generation, we employ a batch size of 16 on 16 H100 GPUs. For text-to-image generation, we set the batch size to 8 with 8 H100 GPUs. All models are trained for 200 steps.

For video generation tasks, we utilize the iStock dataset, where captions are automatically generated from real videos. Specifically, for image-to-video generation, we apply object removal to the first frame of each real video to increase task difficulty and better evaluate the model's generation capability. The training set comprises 10,000 samples, while the test set contains 100 samples. 

For text-to-image generation, we adopt the dataset provided by DanceGRPO for training. The evaluation is performed on the HPSv2 benchmark, which consists of 3,200 prompts.

\end{document}